\pdfoutput=1

\documentclass[11pt]{article}

\usepackage[preprint]{acl}

\usepackage{times}
\usepackage{latexsym}

\usepackage[T1]{fontenc}

\usepackage[utf8]{inputenc}

\usepackage{microtype}

\usepackage{inconsolata}

\usepackage{graphicx}

\usepackage{amsfonts}
\usepackage{amsmath}
\usepackage{booktabs}

\title{Flaw or Artifact? Rethinking Prompt Sensitivity in Evaluating LLMs}

\author{Andong Hua*, Kenan Tang*, Chenhe Gu, Jindong Gu, Eric Wong, Yao Qin}

\author{
  \textbf{Andong Hua\textsuperscript{1 *}},
  \textbf{Kenan Tang\textsuperscript{1 *}},
  \textbf{Chenhe Gu\textsuperscript{2}},
  \textbf{Jindong Gu\textsuperscript{3}},
  \textbf{Eric Wong\textsuperscript{4}},
  \textbf{Yao Qin\textsuperscript{1}}
  \\
  \textsuperscript{1 }UC Santa Barbara,
  \textsuperscript{2 }UC Irvine,
  \textsuperscript{3 }University of Oxford,
  \textsuperscript{4 }University of Pennsylvania
  \\
  \texttt{dongx1997@ucsb.edu, yaoqin@ucsb.edu}
}

\begin{document}
\maketitle
\begin{abstract}
Prompt sensitivity, referring to the phenomenon where paraphrasing (i.e., repeating something written or spoken using different words) leads to significant changes in large language model (LLM) performance, has been widely accepted as a core limitation of LLMs. In this work, we revisit this issue and ask: Is the widely reported high prompt sensitivity truly an inherent weakness of LLMs, or is it largely an artifact of evaluation processes? To answer this question, we systematically evaluate 7 LLMs (e.g., GPT and Gemini family) across 6 benchmarks, including both multiple-choice and open-ended tasks on 12 diverse prompt templates. We find that much of the prompt sensitivity stems from heuristic evaluation methods, including log-likelihood scoring and rigid answer matching, which often overlook semantically correct responses expressed through alternative phrasings, such as synonyms or paraphrases. When we adopt LLM-as-a-Judge evaluations,
we observe a substantial reduction in performance variance and a consistently higher correlation in model rankings across prompts. Our findings suggest that modern LLMs are more robust to prompt templates than previously believed, and that prompt sensitivity may be more an artifact of evaluation than a flaw in the models.

\end{abstract}

\def\thefootnote{*}\footnotetext{Equal contributions.}\def\thefootnote{\arabic{footnote}}

\section{Introduction}
Large Language Models (LLMs) have achieved remarkable success across a wide range of tasks~\cite{hua2024nutribench, lievin2024can}. Moreover, LLMs are good at following diverse instructions, so that users are not required to follow a fixed template when asking a question. This has led to concerns about prompt sensitivity, where differences in prompt phrasing can substantially affect benchmark performance, casting doubt on the reliability of evaluations~\cite{polo2024efficient, mizrahi2024state, chatterjee2024posix, sclar2024quantifying}. More critically, the relative rankings of LLMs can shift substantially depending on the prompt template used~\cite{polo2024efficient, mizrahi2024state}. For example, simply changing the option format from letters (\emph{e.g.}, ``A:'') to numbers (\emph{e.g.}, ``(1)'') completely \emph{reverses} the ranking order of four evaluated open-source models in ARC-Challenge~\cite{arc}. 

Although existing studies have reported that LLMs are highly sensitive to prompt phrasing \cite{voronov2024mind, mizrahi2024state}, this remains counterintuitive given that instruction-tuned LLMs are explicitly optimized to handle a wide range of input formats. For example, instruction-tuning datasets such as FLAN~\cite{longpre2023flan} and Super-NaturalInstructions~\cite{wang2022super} include a diverse collection of tasks (e.g., question answering, summarization, classification) with varying natural language prompt templates~\cite{zhang2023instruction}. This contradiction raises a critical question:

\emph{Is prompt sensitivity an inherent flaw in LLMs, or merely an artifact of the evaluation process?}

\begin{figure*}[t]
    \centering
    \includegraphics[width=1.0\linewidth]{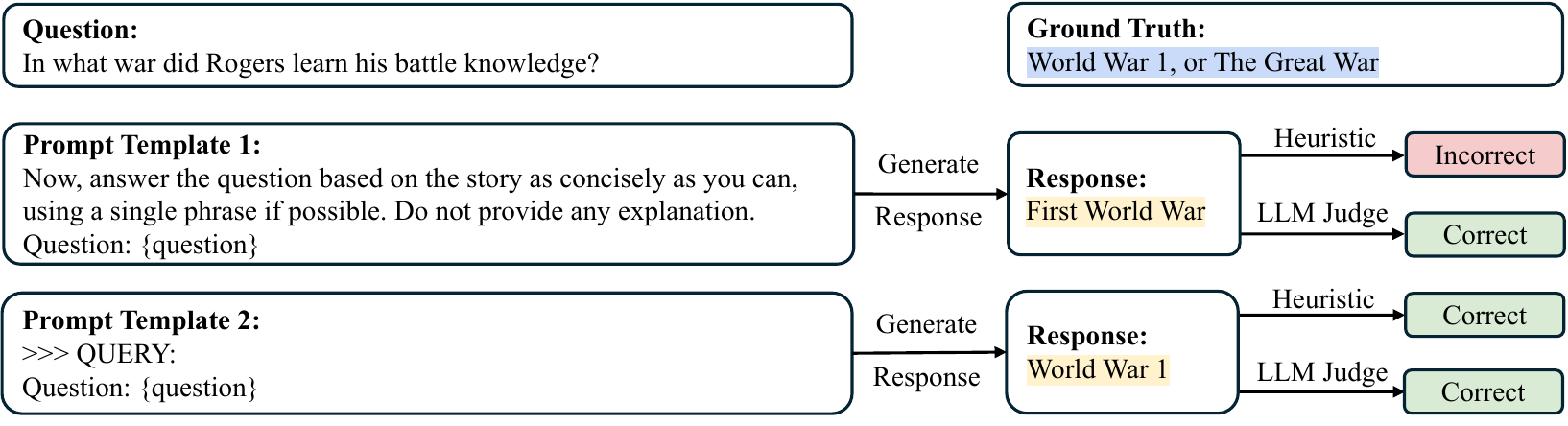}
\caption[LLM response evaluation across prompt templates]{%
When provided with diverse prompt templates, LLMs provide different but semantically equivalent responses. Heuristic evaluation fails to match the different answers with the ground truth, exaggerating prompt sensitivity. In contrast, an LLM judge is able to identify the semantic equivalence consistently.\protect\footnotemark%
}

    \vspace{-1em}
    \label{fig:heuristic_failures}
\end{figure*}

To investigate this, we find that previous studies \cite{voronov2024mind, chatterjee2024posix} typically rely on heuristic evaluation, such as regular-expression-based answer extraction or log-likelihood scoring over candidates. These heuristic evaluation approaches, though historically popular due to their simplicity \cite{zellers-etal-2019-hellaswag, reddy-etal-2019-coqa, brown2020language, mmlu}, may introduce errors when model outputs deviate from expected formats. More specifically, models may generate correct answers, but because their outputs are not aligned with the rigid evaluation format, they are mistakenly marked as incorrect. This issue becomes more pronounced as models have become more open-ended and diverse in their output formats \cite{gpt4o, yang2025qwen3}, potentially leading to inflated estimates of prompt sensitivity (Figure \ref{fig:heuristic_failures}).

\footnotetext{This is an example from NarrativeQA. Heuristic method uses word-level F1; ``Incorrect'' is shown here for illustration purposes, indicating a lower score than a correct answer.}

To rigorously assess whether LLMs truly suffer from prompt sensitivity, we revisit this issue using a more robust evaluation strategy: LLMs as judges. Now widely adopted in recent benchmarks \cite{simpleqa, zheng2023judging}, this approach shifts evaluation from rigid pattern-matching to semantic assessment, enabling a more reliable examination of prompt sensitivity. Compared to heuristics, LLM judges can better handle various output formats, paraphrasing, and ambiguous cases, making them more aligned with human evaluation \cite{liu-etal-2023-g, kocmi-federmann-2023-large, zheng2023judging, wang-etal-2023-chatgpt}.

Using both heuristic methods and LLM-as-a-Judge, we conduct a comprehensive evaluation of four open-source and three closed-source LLMs across 12 prompt templates (without cherry-picking) and six diverse benchmarks, including both multiple-choice and open-ended generation tasks. 
We discover that heuristic evaluation methods often \textbf{exaggerate} the prompt sensitivity of LLMs. For instance, on ARC-Challenge, the performance of Gemma-2.0 varies widely across prompts, with accuracy ranging from 0.25 to 0.90 and a high standard deviation of 0.28. In contrast, when evaluated using LLM-as-a-Judge, its accuracy varies by only 0.17 across the same set of prompt templates, with a much lower standard deviation of just 0.005. Furthermore, the \textit{Spearman rank correlation} measuring model performance rankings across the four open-source models remarkedly improves from 0.31 under heuristic evaluation to 0.92 with LLM-as-a-Judge. These findings suggest that previously reported prompt sensitivity may be significantly overstated due to the limitations of heuristic evaluation. When evaluated with LLM-as-a-Judge, models show far more consistent performance and stable rankings across prompts. Notably, we have conducted a comprehensive human study to verify the reliability of LLM-judges. The high consistency between LLM-judges and human annotators strongly indicates that prompt sensitivity is largely an artifact of the evaluation method rather than an inherent flaw in LLMs.

\section{Method}
Our method consists of three main steps: diverse prompt template construction (Section \ref{sec:diverse-prompt-template-construction}), LLM-as-a-Judge evaluation (Section \ref{sec:llm-as-judge-evaluation}), and prompt template sensitivity measurement (Section \ref{sec:prompt-template-sensitivity-measurement}).

\subsection{Diverse Prompt Template Construction}
\label{sec:diverse-prompt-template-construction}

To evaluate LLM sensitivity to prompt phrasing, we construct diverse prompt templates for each benchmark. These templates vary in instruction wording, answer formatting (e.g., using letters vs.\ numbers), and how responses are requested, while keeping the task content unchanged (Appendix \ref{appendix:prompts}).

In practice, we use GPT-4o to paraphrase the original prompts. For multiple-choice datasets, we create a shared pool of 12 diverse templates used across all benchmarks. For open-ended generation tasks, we generate 12 templates per benchmark to better accommodate domain-specific styles. 

\subsection{LLM-as-a-Judge Evaluation}
\label{sec:llm-as-judge-evaluation}

Heuristic evaluation methods often fail when model outputs deviate from expected formats. To address this limitation, we adopt LLMs as robust judges. In this approach, an LLM judge is given the original question, the correct answer, and the model's predicted response. The judge is prompted to determine whether the predicted response semantically matches the correct answer (Appendix~\ref{appendix:prompts-for-llm-as-judge}). 

While the overall format remains consistent, we introduce minor benchmark-specific adjustments to the judging prompt. For example, for GPQA, the judge is instructed to “Ignore all explanation” to ensure it focuses solely on the final answer.
\vspace{-2mm}
\subsection{Prompt Sensitivity Measurement}
\label{sec:prompt-template-sensitivity-measurement}

We measure sensitivity with two metrics: \textit{performance variation} and \textit{ranking consistency}.

\paragraph{Performance Variation.} For each model and dataset, we compute the accuracy under every prompt template and report the standard deviation across all prompt variants. Let \( P = \{p_1, p_2, \dots, p_n\} \) denote the set of prompt templates for a given benchmark \( D \), and let \( f \) be the model under evaluation. The performance of model \( f \) under prompt \( p_i \) is denoted by \( A_{f,D}^{p_i} \). The prompt sensitivity of model \( f \) on dataset \( D \) is then quantified as:
$\mathrm{std}_f = \mathrm{StdDev} \left( \left\{ A_{f,D}^{p_i} \right\}_{i=1}^n \right)$.
A lower standard deviation indicates that the model's performance is stable across different prompt templates.

\paragraph{Ranking Consistency.} 
\label{sec:ranking_consistency}
Beyond absolute performance, we also measure how model rankings vary across prompt templates. Given a set of \( K \) models, we rank them based on their performance under each prompt and compute pairwise \textit{Spearman's rank correlation} between all prompt pairs~\cite{spearman1904}. Given two templates \( p_i \) and \( p_j \), and the corresponding performance vectors \(\{A_{f_k,D}^{p_i}\}_{k=1}^K\) and \(\{A_{f_k,D}^{p_j}\}_{k=1}^K\), we calculate Spearman's rank correlation coefficient $\rho_{ij} = 1 - \frac{6 \sum_{k=1}^K d_k^2}{K(K^2 - 1)}$, 
where \( d_k \) is the difference in rankings of the \(k\)-th model under prompts \(p_i\) and \(p_j\), and \(K\) is the number of models. The rank correlation coefficient, \( \rho \), ranges from \(-1\) to \(1\), with higher values indicating stronger agreement in ranking consistency.

To measure overall ranking consistency, we compute the mean Spearman's rank correlation coefficient, denoted as \( \bar{\rho} \), across all pairs of prompt templates. This mean score \( \bar{\rho} \) serves as a comprehensive metric for evaluating the stability of model rankings under prompt variation. A higher \( \bar{\rho} \) suggests that evaluations are more robust and less dependent on the specific prompt phrasing.

\section{Results and Discussion}
\subsection{Experimental Setup}

\paragraph{Models.} We evaluate LLaMA-3.1-8B-Instruct (LLaMA-3.1)~\cite{llama3}, Qwen2-7B-Instruct (Qwen-2)~\cite{qwen2}, Gemma-2-9B-it (Gemma-2)~\cite{gemma2}, Ministral-8B-Instruct-v0.2 (Ministral)~\cite{ministral8b}, GPT-4o-mini (July 2024), GPT-4.1-mini (April 2025), and Gemini 2.0 Flash (February 2025).

\paragraph{Benchmarks.}
We evaluate on six benchmarks covering both multiple-choice and open-ended tasks.
The multiple-choice datasets include ARC-Challenge~\cite{arc}, GPQA-diamond~\cite{gpqa}, and OpenbookQA~\cite{openbookqa}, where answers are selected from discrete options (e.g., A/B/C/D). For these tasks, heuristic evaluation uses log-likelihood scoring over answer options. The open-ended datasets include NarrativeQA~\cite{narrativeqa}, MATH~\cite{hendrycks2021measuring}, and SimpleQA~\cite{simpleqa}, where model responses are free-form. For NarrativeQA and MATH, heuristic evaluation applies format-specific extraction and normalization (see Appendix~\ref{appendix:heuristic}). For SimpleQA, no rule-based parser is available, so we report results only under the LLM-as-a-Judge framework. All evaluations use greedy decoding to ensure deterministic outputs.

\paragraph{LLM-as-a-Judge.} Unless otherwise specified, we employ Gemini 2.0 Flash as the LLM-as-a-Judge across all benchmarks.

\begin{figure*}[htbp]
    \centering
    \includegraphics[width=0.32\linewidth]{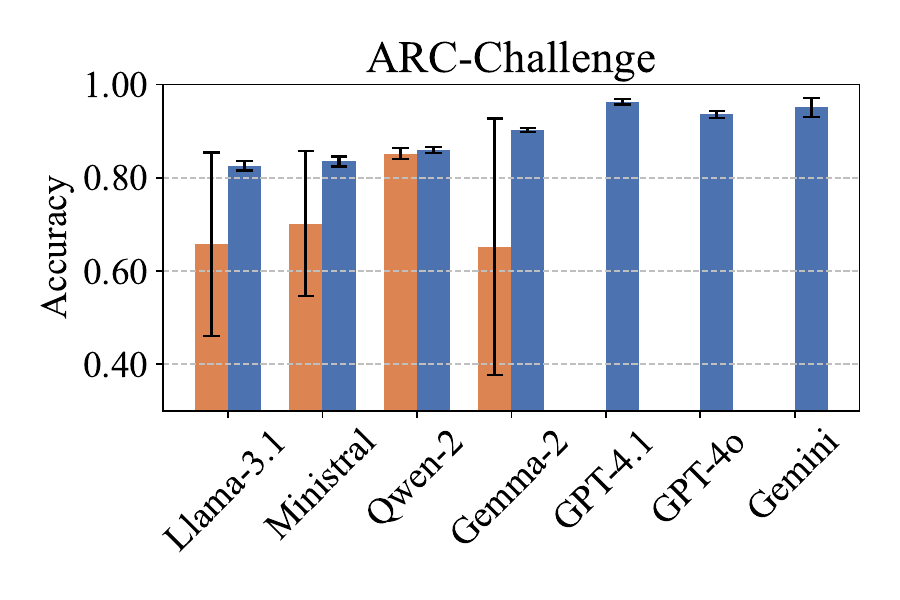}
    \includegraphics[width=0.32\linewidth]{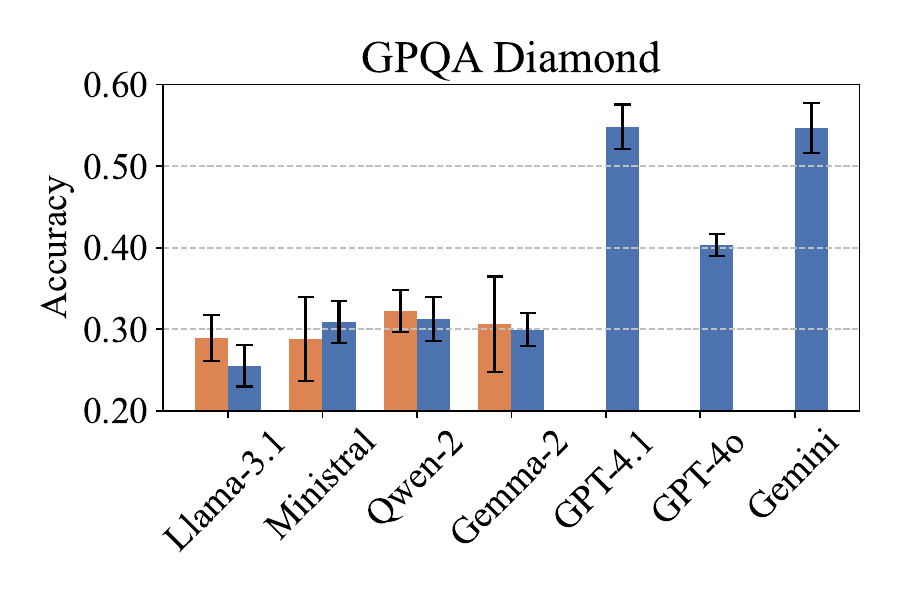}
    \includegraphics[width=0.32\linewidth]{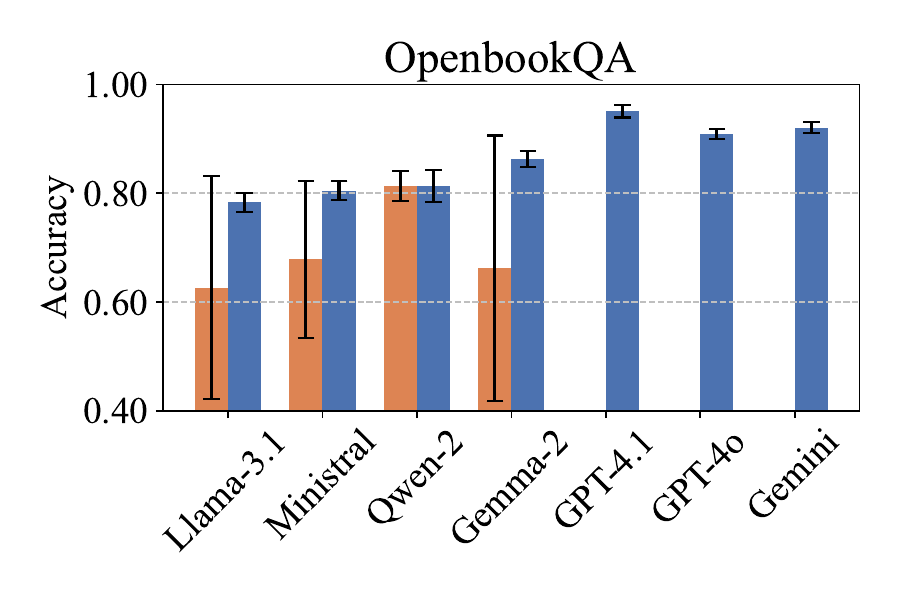}
    \includegraphics[width=0.32\linewidth]{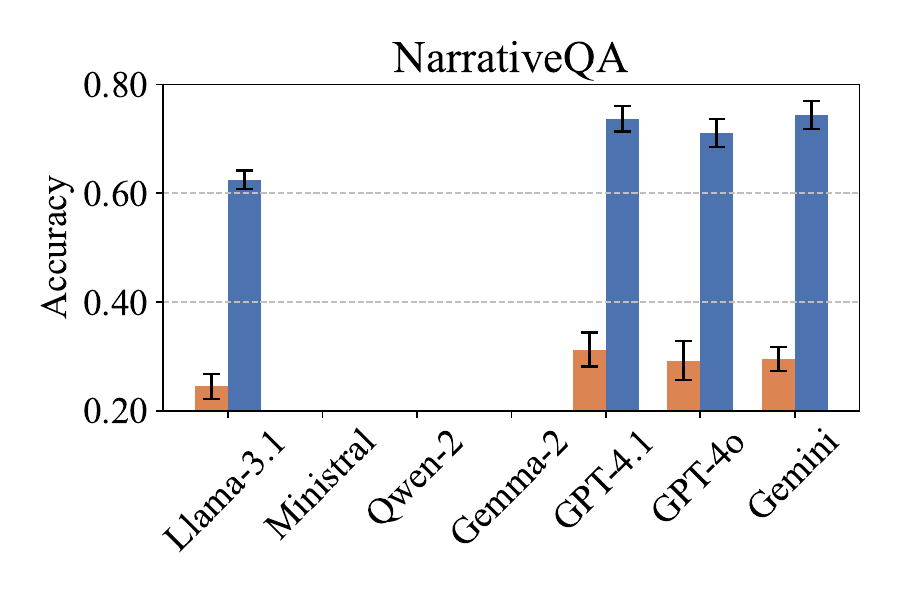}
    \includegraphics[width=0.32\linewidth]{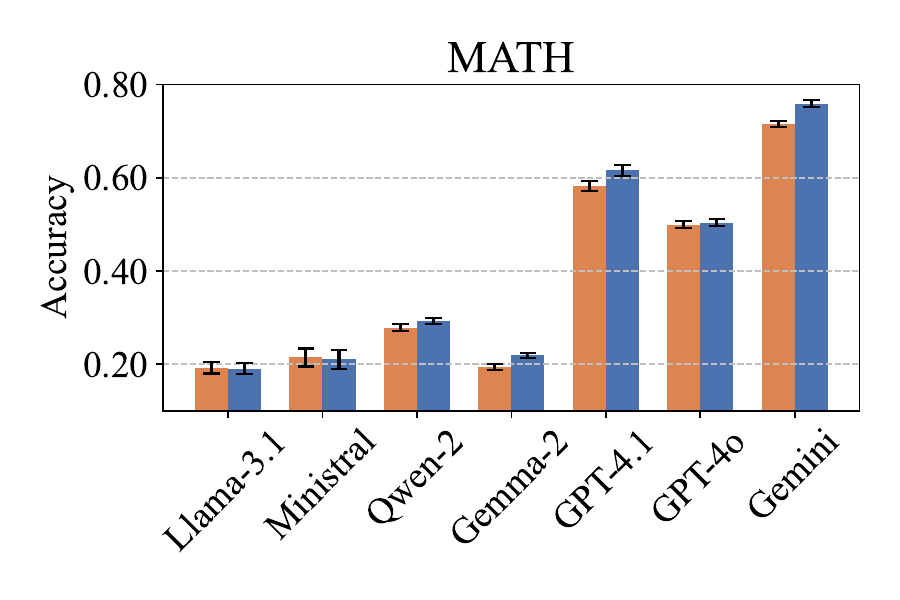}
    \includegraphics[width=0.32\linewidth]{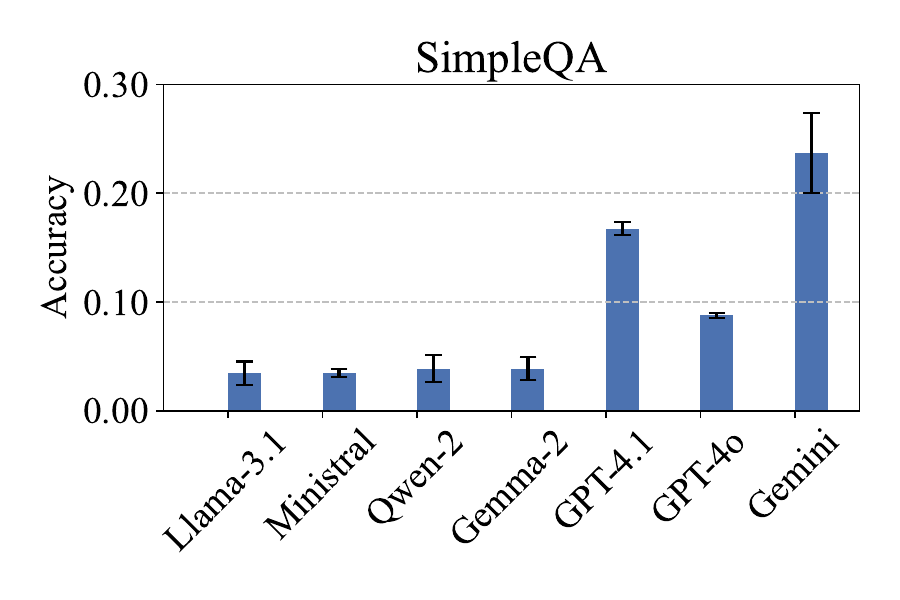}
    \includegraphics[width=0.35\linewidth]{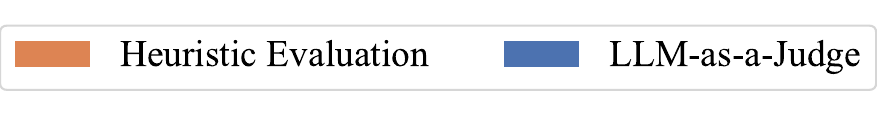}
    \vspace{-1em} 
    \caption{The mean and standard deviation of performance across different prompt templates. For all 6 datasets, we show the statistics for all pairs of evaluation methods and models, excluding the cases when the model's context length is not enough for the task or when the heuristic evaluation method is not available. The standard deviation of the LLM-as-a-Judge method is always low. For NarrativeQA, the absence of results for Mistral and Qwen is primarily due to the long-context requirement of this dataset.}

    \vspace{-1em}
    \label{fig:acc-std}
\end{figure*}

\begin{table}[h]
    \centering
    \small
        \begin{tabular}{ccc}
        \toprule
        Dataset & $\bar\rho_{\text{Heuristic}}$ & $\bar{\rho}_{\text{LLM}}$ \\
        \midrule
        ARC-Challenge\textsuperscript{*} & 0.3036 & 0.9546 (0.9187) \\
        OpenbookQA\textsuperscript{*}    & 0.4212 & 0.9386 (0.7360) \\
        GPQA Diamond\textsuperscript{*}  & 0.1542 & 0.8960 (0.5048) \\
        NarrativeQA\textsuperscript{†}   & 0.5927 & 0.8662 \\
        MATH                             & 0.9593 & 0.9647 \\
        SimpleQA                         & --     & 0.8121 \\
        \bottomrule
        \end{tabular}
        
\caption{
Average Spearman rank correlation ($\bar{\rho}$) across prompt templates using heuristic vs.\ LLM-as-a-Judge. 
Results use all 7 models unless noted: \textsuperscript{*} heuristic uses only 4 open-source models (parentheses show LLM-as-a-Judge restricted to the same 4 models); 
\textsuperscript{†} NarrativeQA uses LLaMA-3.1 and 3 proprietary models due to context limits; 
``--'' indicates heuristic not applicable. 
}

    \vspace{-1em}
    \label{tab:rank-correlation}
\end{table}

\subsection{Heuristic Evaluation Exaggerates Prompt Sensitivity of LLMs}

When comparing the performance of the model under heuristic evaluation and LLM-as-a-Judge, we find that the heuristic methods exhibit significantly greater sensitivity to prompt variation (Figure~\ref{fig:acc-std}). On ARC-Challenge, all open-source models except Qwen-2 show much higher standard deviations under heuristics. For instance, Gemma-2.0 yields a deviation of 0.28, versus just 0.005 with LLM-as-a-Judge. Its accuracy range spans 0.25–0.90 under heuristics, compared to only 0.17 with LLM-as-a-Judge. Additionally, mean accuracy improves under LLM-as-a-Judge, suggesting heuristic methods often miss valid answers due to overly rigid extraction rules.

Beyond variance in accuracy, we also assess ranking consistency across prompts (Section~\ref{sec:ranking_consistency}). On ARC-Challenge, the average Spearman rank correlation across prompts among open-source models increases from 0.30 (heuristics) to 0.92 (LLM-as-a-Judge), and further to 0.95 when proprietary models are included. On NarrativeQA, the correlation rises from 0.40 (heuristics) to 0.87 (LLM-as-a-Judge). 
These findings suggest that prompt sensitivity observed in prior work is largely an artifact of heuristic evaluation, not an inherent flaw of LLMs.

\paragraph{Well-designed heuristic methods show low prompt sensitivity similar to LLM-as-a-Judge.}
For MATH~\cite{hendrycks2021measuring}, the heuristic approach incorporates symbolic simplification, expression normalization, and equivalence checking using tools such as \texttt{sympy}. Under these conditions, we observe prompt sensitivity results that are comparable to those obtained via LLM-as-a-Judge evaluation, with similarly low accuracy variance and high ranking consistency. These results indicate that, with sufficient domain-specific prompt engineering, heuristic methods can provide stable evaluations. This further supports the conclusion that modern LLMs exhibit less prompt-induced variance than previously reported.

\paragraph{Newly proposed benchmarks exhibit low prompt sensitivity.}
We extend the evaluation to SimpleQA~\cite{simpleqa}, a newly proposed benchmark for factual and commonsense reasoning. As no official heuristic evaluation is available, we use LLM-as-a-Judge by default. Applying our prompt sensitivity analysis, we observe a low standard deviation and a high Spearman rank correlation of 0.8121 across 12 prompt templates. These results indicate that model performance remains stable across different prompt variations.

\subsection{Consistency of LLM-as-a-Judge Evaluation Across LLM Judges}

To further examine whether evaluation outcomes vary significantly when using different LLMs as judges, 
we conducted an additional analysis on the ARC-Challenge dataset with GPT-4o-mini as the judge. 
As shown in Table~\ref{tab:judge-consistency}, both the standard deviation across different prompts 
and the ranking correlation remain consistent regardless of whether GPT or Gemini is used as the judge. 
This suggests that LLM-as-a-Judge evaluations are robust across different LLMs, reinforcing our main argument 
that prompt sensitivity is largely a byproduct of heuristic evaluation rather than true model instability.

\begin{table}[t]
    \centering
    \small
    \begin{tabular}{lcc}
        \toprule
        Model & GPT Judge & Gemini Judge \\
        \midrule
        Llama-3.1 & 0.8276 $\pm$ 0.0094 & 0.8257 $\pm$ 0.0102 \\
        Mistral   & 0.8349 $\pm$ 0.0105 & 0.8349 $\pm$ 0.0107 \\
        Qwen2     & 0.8590 $\pm$ 0.0063 & 0.8592 $\pm$ 0.0059 \\
        Gemma-2   & 0.9027 $\pm$ 0.0048 & 0.9027 $\pm$ 0.0046 \\
        Gemini    & 0.9590 $\pm$ 0.0050 & 0.9512 $\pm$ 0.0200 \\
        GPT-4.1   & 0.9617 $\pm$ 0.0037 & 0.9626 $\pm$ 0.0054 \\
        GPT-4o    & 0.9371 $\pm$ 0.0050 & 0.9360 $\pm$ 0.0069 \\
        \midrule
        Rank Correlation & 0.9621 & 0.9546 \\
        \bottomrule
    \end{tabular}
    
    \caption{
    Accuracy scores (Mean $\pm$ Std) across prompts in ARC-Challenge with different LLM judges. 
    Results show high consistency between GPT and Gemini judges.
    }
    \vspace{-1em}
    \label{tab:judge-consistency}
\end{table}

\section{LLM-as-a-Judge Evaluation Aligns with Human Annotations}
\label{sec:human-annotations}

To assess the reliability of LLM-judges, we compare them against human annotations. We recruit human annotators to manually evaluate answer correctness. For each dataset, we randomly sample 50 questions and collect answers from one model under all 12 prompt templates, yielding 600 answers per dataset-model pair. For ARC-Challenge, OpenbookQA, and GPQA Diamond, we evaluate Gemma-2; for NarrativeQA, MATH, and SimpleQA, we evaluate GPT-4.1-mini. We also report results on the combined set of all six subsets.

Human annotators evaluate whether each answer matches the corresponding ground truth. Their judgments are highly consistent, and majority voting is used to resolve the few cases of disagreement. Moreover, the majority-voted outcomes closely align with the LLM-judge results, underscoring the reliability of LLM-as-a-Judge evaluation. More details and the original human annotation instructions can be found in Appendix \ref{app:human-details}. 

From the results, we draw two key observations.

\paragraph{Observation 1} Human-annotated results show consistently high agreement on the answer correctness (Table \ref{tab:prompt-template-agreement}), with a high Fleiss' $\kappa$ over 0.6 \cite{fleiss1971measuring}. This provides strong evidence that the answer correctness does not vary substantially across different prompt templates.

\paragraph{Observation 2} The rate of perfect agreement is also notably high (Table \ref{tab:prompt-template-agreement}). We define perfect agreement as cases where the answer correctness is identical across all prompt templates for a given question. This further reinforces our conclusion that prompt templates have minimal impact on the answer correctness.

\begin{table}[h]
    \centering
    \small
        \begin{tabular}{ccc}
        \toprule
        Dataset & Agreement & Perfect Agreement\\
        \midrule
        Arc-Challenge & 0.7654 & 86\%\\
        OpenbookQA    & 0.7250 & 80\%\\
        GPQA-Diamond  & 0.6708 & 52\%\\
        NarrativeQA   & 0.7258 & 66\%\\
        MATH          & 0.7859 & 68\%\\
        SimpleQA      & 0.7881 & 88\%\\
        \midrule
        Combined      & 0.8132 & 73\%\\
        \bottomrule
        \end{tabular}
        
    \caption{
    Correctness of answers shows minimal variation across prompt templates, based on human annotations. ``Agreement'' is measured by Fleiss’ $\kappa$ \cite{fleiss1971measuring} across 12 prompt templates (higher is better). ``Perfect Agreement'' is defined as cases where all 12 prompt templates yield the same correctness judgment for a given question. We report the percentage of questions with perfect agreement.
    }
    \vspace{-1em}
    \label{tab:prompt-template-agreement}
\end{table}

\section{Related Work}

The ranking inconsistency with diverse prompt templates has been widely reported \cite{polo2024efficient,mizrahi2024state,chatterjee2024posix,sclar2024quantifying}. \citet{mizrahi2024state} conducted a large-scale study showing significant accuracy differences across prompt variants. \citet{voronov2024mind} further showed that no prompt format consistently performs best across models. To address this, prior work often assumes that LLMs are inherently unstable to prompt changes. For example, \citet{polo2024efficient} estimates the distribution of accuracy across prompts to improve evaluation efficiency. However, all existing methods attribute the sensitivity to model behavior. In contrast, we show that a key factor is the heuristic evaluation protocol itself, which often leads to misclassification of correct outputs and overstates prompt sensitivity.

\section{Conclusion}

In this work, we demonstrate that much of the observed prompt sensitivity in LLM evaluations is not due to inherent model weaknesses, but rather an artifact introduced by heuristic evaluation methods. Through comprehensive experiments using LLM-as-a-Judge across multiple benchmarks and prompt templates, we reveal that model performance and rankings are substantially more stable and reliable than previously reported.  We hope this work sheds light on prompt sensitivity in LLM evaluation and encourages broader adoption of LLM-as-a-Judge to evaluate the true capabilities of LLMs.

\section*{Limitations}

Due to computational constraints, we evaluate each benchmark using only 12 prompt templates. However, we find that results are stable across scales. For example, on ARC-Challenge, the ranking consistency and variance metrics using 12 prompts closely match those obtained with over 100 prompts, suggesting that our analysis is representative.

\bibliography{custom}

\appendix

\section{Diverse Prompts}
\label{appendix:prompts}

In this section, we list the diverse prompt templates we use for each benchmark.

For ARC-Challenge, GPQA, and OpenbookQA, we use the following 12 prompt templates:
\begin{raggedright}
\begin{enumerate}
    \item Evaluate the choices and select the most appropriate answer.\textbackslash n\{question\}\textbackslash nThe options are as follows:\textbackslash nOption A: \{first\_option\}\textbackslash nOption B: \{second\_option\}\textbackslash nOption C: \{third\_option\}\textbackslash nOption D: \{fourth\_option\}\textbackslash nYour answer should be formatted as:\textbackslash n'I have chosen option [choice]'\textbackslash nwhere [choice] is your selected answer.\textbackslash n
    \item Review the available options and select the one you think is correct.\{question\}\textbackslash nAvailable answers include:\textbackslash nA ) \{first\_option\}\textbackslash nB ) \{second\_option\}\textbackslash nC ) \{third\_option\}\textbackslash nD ) \{fourth\_option\}\textbackslash n\textbackslash nResponse: 
    \item Select the correct answer based on your understanding.\textbackslash n\{question\}\textbackslash nPick from the following options:\textbackslash n[A] \{first\_option\}\textbackslash n[B] \{second\_option\}\textbackslash n[C] \{third\_option\}\textbackslash n[D] \{fourth\_option\}\textbackslash nPlease respond with 'Option [choice]'.\textbackslash n
    \item Evaluate the options presented and select the most suitable.\{question\}\textbackslash nAvailable answers:\textbackslash n[A] \{first\_option\}\textbackslash n[B] \{second\_option\}\textbackslash n[C] \{third\_option\}\textbackslash n[D] \{fourth\_option\}\textbackslash n\textbackslash nExpress your choice as: 'The answer is [choice].'\textbackslash nwhere [choice] is your selected option.\textbackslash n
    \item Based on the question presented, choose the most fitting response.\{question\}\textbackslash nAvailable answers are:\textbackslash nA: \{first\_option\}\textbackslash nB: \{second\_option\}\textbackslash nC: \{third\_option\}\textbackslash nD: \{fourth\_option\}\textbackslash nPlease provide your response in the following format:\textbackslash n'Your choice: [option]'\textbackslash nwhere [option] corresponds to the letter or number you selected.\textbackslash n
    \item From the options below, select the response that you believe is correct.\{question\}\textbackslash nChoices to consider:\textbackslash n1. \{first\_option\}\textbackslash n2. \{second\_option\}\textbackslash n3. \{third\_option\}\textbackslash n4. \{fourth\_option\}\textbackslash nResponse: 
    \item Select your answer from the provided list of options.\textbackslash n\{question\}\textbackslash nOptions are:\textbackslash nThe choice is A: \{first\_option\}\textbackslash nThe choice is B: \{second\_option\}\textbackslash nThe choice is C: \{third\_option\}\textbackslash nThe choice is D: \{fourth\_option\}\textbackslash n\textbackslash nChoose your answer: 
    \item After considering the options, choose the best possible answer.\{question\}\textbackslash nThe following choices are available:\textbackslash nA: \{first\_option\}\textbackslash nB: \{second\_option\}\textbackslash nC: \{third\_option\}\textbackslash nD: \{fourth\_option\}\textbackslash nState your answer as:\textbackslash n'Answer: [choice]'\textbackslash n
    \item Analyze the selections and provide your choice.\textbackslash n\{question\}\textbackslash nYour options are listed below:\textbackslash nOption 1 - \{first\_option\}\textbackslash nOption 2 - \{second\_option\}\textbackslash nOption 3 - \{third\_option\}\textbackslash nOption 4 - \{fourth\_option\}\textbackslash n\textbackslash nYour response: 
    \item Consider the following question and determine the right response.\textbackslash n\{question\}\textbackslash nWhich of the following answers do you prefer?\textbackslash nOption 1: \{first\_option\}\textbackslash nOption 2: \{second\_option\}\textbackslash nOption 3: \{third\_option\}\textbackslash nOption 4: \{fourth\_option\}\textbackslash nI select: 
    \item Determine which option best answers the question asked.\textbackslash n\{question\}\textbackslash nPossible choices are as follows:\textbackslash nOption [A] \{first\_option\}\textbackslash nOption [B] \{second\_option\}\textbackslash nOption [C] \{third\_option\}\textbackslash nOption [D] \{fourth\_option\}\textbackslash n\textbackslash nFinal answer: 
    \item Identify the option that best answers the question posed.\{question\}\textbackslash nConsider these choices:\textbackslash nSelect option 1: \{first\_option\}\textbackslash nSelect option 2: \{second\_option\}\textbackslash nSelect option 3: \{third\_option\}\textbackslash nSelect option 4: \{fourth\_option\}\textbackslash n\textbackslash nChoice provided: 
\end{enumerate}
\end{raggedright}
In the prompt templates, \{question\} is the question, and \{first\_option\}, \{second\_option\}, \{third\_option\}, and \{fourth\_option\} are the options.

For NarrativeQA, we use the following 12 prompt templates:
\begin{raggedright}
\begin{enumerate}
    \item You are given a story, which can be either a novel or a movie script, and a question. Answer the question asconcisely as you can, using a single phrase if possible. Do not provide any explanation.\textbackslash n\textbackslash n Story: \{context\}\textbackslash n\textbackslash n Now, answer the question based on the story as concisely as you can, using a single phrase if possible. Do not provide any explanation.\textbackslash n\textbackslash n Question: \{question\}\textbackslash n\textbackslash n Answer:
    \item Below is an excerpt from a mystery or thriller story, followed by a question. Provide the most accurate answer you can in a single phrase or sentence fragment. No elaboration is needed.\textbackslash n\textbackslash nStory: \{context\}\textbackslash n\textbackslash nExamine the situation carefully and respond.\textbackslash n\textbackslash nQuestion: \{question\}\textbackslash n\textbackslash nAnswer:
    \item You are presented with a passage from literary fiction or cinematic writing and a comprehension question. Respond succinctly with a phrase. Avoid any additional commentary.\textbackslash n\textbackslash nStory: \{context\}\textbackslash n\textbackslash nAnalyze and respond concisely.\textbackslash n\textbackslash nQuestion: \{question\}\textbackslash n\textbackslash nAnswer:
    \item A tale from a distant world or magical land is told below, followed by a question from a curious scholar. Give your answer using only a few words. No need to explain the lore.\textbackslash n\textbackslash nStory: \{context\}\textbackslash n\textbackslash nWhat say you?\textbackslash n\textbackslash nQuestion: \{question\}\textbackslash n\textbackslash nAnswer:
    \item You're reading a gritty tale from the backstreets of the city. A question follows. Keep your answer clipped, clean, and under the radar—just a phrase, no fluff.\textbackslash n\textbackslash nStory: \{context\}\textbackslash n\textbackslash nHere's the case:\textbackslash n\textbackslash nQuestion: \{question\}\textbackslash n\textbackslash nAnswer:
    \item Welcome to *Plot Points*! We’ll give you a story snippet and a question—your job is to give the fastest, most precise answer possible. One phrase, no lifelines!\textbackslash n\textbackslash nStory: \{context\}\textbackslash n\textbackslash nLet’s play!\textbackslash n\textbackslash nQuestion: \{question\}\textbackslash n\textbackslash nAnswer:
    \item The record shows the following account. A question will now be entered into the record. Provide your answer in a short, factual phrase. No commentary permitted.\textbackslash n\textbackslash nStory: \{context\}\textbackslash n\textbackslash nDeposition Question:\textbackslash n\textbackslash nQuestion: \{question\}\textbackslash n\textbackslash nAnswer:
    \item Read the excerpt. Answer the question. Keep it short.\textbackslash n\textbackslash nStory: \{context\}\textbackslash n\textbackslash nQuestion: \{question\}\textbackslash n\textbackslash nAnswer:
    \item Accessing archive… Story fragment retrieved from Galactic Chronicles. A query follows. Respond with the most relevant concept or phrase. Do not explain.\textbackslash n\textbackslash nStory: \{context\}\textbackslash n\textbackslash n>{}>{}> QUERY:\textbackslash n\textbackslash nQuestion: \{question\}\textbackslash n\textbackslash n>{}>{}> RESPONSE:\textbackslash n\textbackslash nAnswer:
    \item Once upon a time, a story was told. Now a little question is asked. Answer it kindly and briefly—just a few words will do. No need to explain why.\textbackslash n\textbackslash nStory: \{context\}\textbackslash n\textbackslash nHere comes the question:\textbackslash n\textbackslash nQuestion: \{question\}\textbackslash n\textbackslash nAnswer:
    \item From the folds of a lyrical tale, a question emerges like morning light. Respond with a single phrase, a shard of truth—no more, no less.\textbackslash n\textbackslash nStory: \{context\}\textbackslash n\textbackslash nWhisper your reply:\textbackslash n\textbackslash nQuestion: \{question\}\textbackslash n\textbackslash nAnswer:
    \item Intel received. Narrative extracted. Stand by for situational query. Your task: deliver the answer in minimal terms. Do not elaborate.\textbackslash n\textbackslash nStory: \{context\}\textbackslash n\textbackslash nMission Query:\textbackslash n\textbackslash nQuestion: \{question\}\textbackslash n\textbackslash nAnswer:
\end{enumerate}
\end{raggedright}
In the prompt templates, \{context\} is the context, and \{question\} is the question.

For MATH, we use the prompt template \{text1\}\{question\}\{text2\}, where \{text1\} and \{text2\} are two strings that enclose the question. The following 12 pairs of (\{text1\}, \{text2\}) are used:
\begin{enumerate}
    \item (\{empty\_string\}, \textbackslash nAnswer:\textbackslash n)
    \item (Problem::\textbackslash n, \textbackslash nAnswer:\textbackslash n)
    \item (Problem::\textbackslash n, \textbackslash nAnswer:\textbackslash n)
    \item (Task:\textbackslash n\textbackslash n, \textbackslash n\textbackslash nSolution:)
    \item (Solve the following math problem:\textbackslash n\textbackslash n, \textbackslash nAnswer:\textbackslash n)
    \item (Solve the following math problem:\textbackslash n\textbackslash n, \textbackslash nAnswer:\textbackslash n)
    \item (**Problem Statement**:\textbackslash n\textbackslash n, \textbackslash n\textbackslash nSolution:)
    \item (Problem::\textbackslash n, \textbackslash n\textbackslash nSolution:)
    \item (Solve the following math problem:\textbackslash n\textbackslash n, \textbackslash n\textbackslash nSolution:)
    \item (**Problem Statement**:\textbackslash n\textbackslash n, \textbackslash n\textbackslash nSolution:)
    \item (**Problem Statement**:\textbackslash n\textbackslash n, \textbackslash nAnswer:\textbackslash n)
    \item (\{empty\_string\}, \textbackslash nAnswer:\textbackslash n)
\end{enumerate}
Since MATH uses few-shot prompting for evaluation, we further change the examples provided for each prompt template. Hence, while two pairs of (\{text1\}, \{text2\}) could be the same, the actual prompt template is different.

For SimpleQA, we use the prompt template \{instruction\}\{question\}, where \{question\} is the original questions in the benchmark, and \{instruction\} is one of the 12 following strings:
\begin{enumerate}
    \item \{empty\_string\}
    \item Ready your reasoning—consider the challenge that follows.\textbackslash n\textbackslash n
    \item Take a thoughtful pause, then craft your best response to the prompt beneath this line.\textbackslash n\textbackslash n
    \item Showcase your insight by addressing the upcoming question.\textbackslash n\textbackslash n
    \item Put your analytical lens on and dive into the inquiry below.\textbackslash n\textbackslash n
    \item Channel your inner detective: examine the next question and present your findings.\textbackslash n\textbackslash n
    \item Let your knowledge shine—respond thoughtfully to the statement that follows.\textbackslash n\textbackslash n
    \item Engage your critical thinking skills and tackle the question that appears next.\textbackslash n\textbackslash n
    \item Apply the concepts you've mastered to answer the forthcoming inquiry.\textbackslash n\textbackslash n
    \item Use evidence and reasoning to construct your answer to the question below.\textbackslash n\textbackslash n
    \item Approach the next problem with curiosity and craft a clear solution.\textbackslash n\textbackslash n
    \item Demonstrate what you've learned by addressing the prompt that follows.\textbackslash n\textbackslash n
\end{enumerate}

\section{Prompts for LLM-as-a-Judge}
\label{appendix:prompts-for-llm-as-judge}
\begin{figure}
    \centering
    \includegraphics[width=1.0\linewidth]{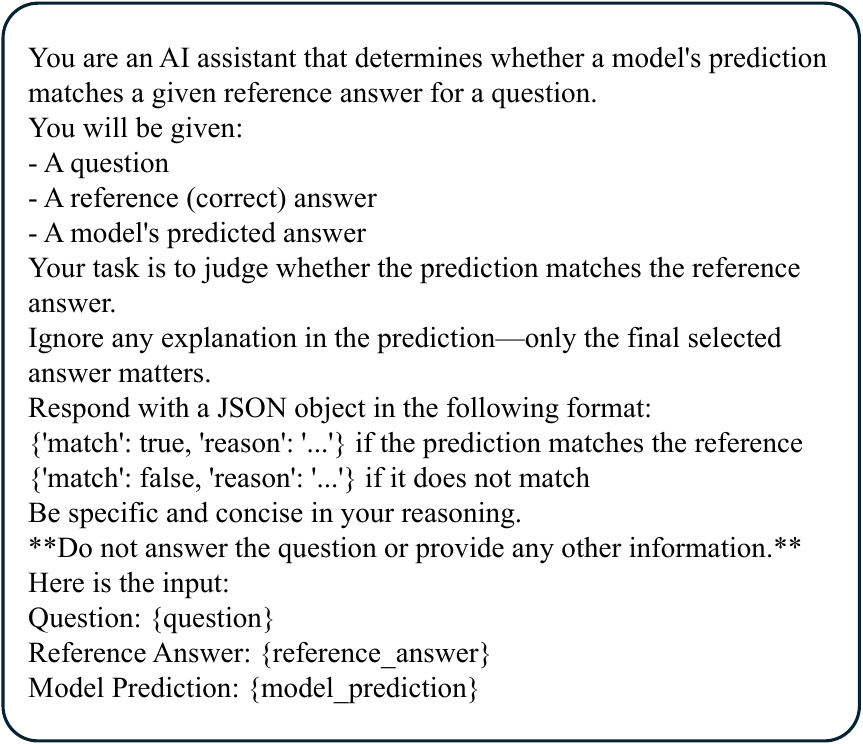}
    \caption{An example of a judging prompt. After filling in the question, reference answer, and model prediction, we send the prompt to an LLM judge to get the result.}
    \label{fig:llm_judge_prompt}
\end{figure}

Figure~\ref{fig:llm_judge_prompt} shows the prompt we use for LLM-as-a-Judge. For each benchmark, we make minor task-specific modifications to the judging prompt. For SimpleQA, we use the official prompt from OpenAI.

\section{Heuristic Evaluation Details}
\label{appendix:heuristic}

\paragraph{NarrativeQA.}
Following LongBench~\cite{bai2024longbench}, we compute word-level F1 overlap between normalized predictions and references. Normalization includes lowercasing, removing punctuation and articles (a, an, the), and collapsing whitespace. For example, the prediction “Fifty years” and the reference “50 years” would result in a partial overlap and an F1 score of 0.5.

\paragraph{MATH.}
Heuristic evaluation typically extracts the final answer from LaTeX-formatted expressions such as \verb|\boxed{...}| or \verb|\fbox{...}|. This method assumes the model explicitly marks its answer with these delimiters. We follow this approach in our experiments. See Appendix~\ref{appendix:prompts} for examples.

\paragraph{SimpleQA.}
No official heuristic parser is currently available, making rule-based evaluation infeasible. We therefore rely solely on the LLM-as-a-Judge method for this dataset.

\section{Human Annotation Details}
\label{app:human-details}

We recruit three undergraduate students from UCSB with no prior exposure to this study. The undergraduate students are Asian or Asian American, and all of them are fluent in English. In total, we collected $6 \times 50 \times 12 \times 3 = 10,800$ human annotations across different datasets and models. Each annotator spends approximately six hours on the task and receives a 
\$75 Amazon gift card, which is the maximum compensation permitted by the department. The annotators are informed that their work will be included in a research paper, but the topic of the paper is not disclosed in order to minimize potential bias.

Inter-annotator agreement is high (Table \ref{tab:human-annotation-agreement}), indicating the overall quality and consistency of human annotations. In the few cases of disagreement, we apply majority voting to determine the final label.

Furthermore, human annotations (after majority voting) exhibit strong agreement with the LLM-judge (Table \ref{tab:human-annotation-agreement}). This demonstrates that the LLM-judge provides reliable correctness judgments when comparing model answers against ground truth across diverse prompt templates.

\begin{table}[h]
    \centering
    \small
        \begin{tabular}{ccc}
        \toprule
        Dataset & Human-Human & Human-LLM\\
        \midrule
        Arc-Challenge & 0.9856 & 0.9785\\
        OpenbookQA    & 0.9919 & 1.0000\\
        GPQA-Diamond  & 0.9083 & 0.9778\\
        NarrativeQA   & 0.6870 & 0.6699\\
        MATH          & 0.8943 & 0.8563\\
        SimpleQA      & 0.7812 & 0.9849\\
        \midrule
        Combined      & 0.9010 & 0.9247\\
        \bottomrule
        \end{tabular}
        
    \caption{
    Human annotation results show strong consistency across annotators. After majority voting, they also align closely with LLM-as-a-Judge evaluations. We report Fleiss’ $\kappa$ \cite{fleiss1971measuring} for agreement among three human annotators (Human–Human) and Cohen’s $\kappa$ \cite{cohen1960coefficient} for agreement between majority-voted human annotations and LLM-as-a-Judge results (Human–LLM).
    }
    \vspace{-1em}
    \label{tab:human-annotation-agreement}
\end{table}

The human annotation instructions we use are shown below (from ``Hi'' to ``[GOOGLE SHEETS LINK]'').
\\\\
\noindent Hi [NAME],

\noindent I have shared a Google spreadsheet with you. Your task would be comparing the model's answer (``pred'' column) with the ground truth (``gt'' column), given the question (``question'' column). Put 1 in the ``human\_evaluation'' column if the model's answer is correct. Put 0 in the ``human\_evaluation'' column if the model's answer is wrong. If the model's answer is ambiguous, put 0.

\noindent [GOOGLE SHEETS LINK]

\section{Additional Results on Older LLMs}
To examine whether prompt sensitivity is a persistent issue across model generations, we further evaluated two earlier instruction-tuned LLMs: \textbf{Llama-2-7B-Chat} and \textbf{Mistral-7B-Instruct-v0.1}. 

\begin{table}[t]
    \centering
    \small
    \begin{tabular}{lcc}
        \toprule
        & Llama-2 & Mistral \\
        \midrule
        Heuristic Mean      & 0.3706 & 0.6125 \\
        Heuristic Std       & 0.1413 & 0.0389 \\
        LLM-as-a-Judge Mean   & 0.5587 & 0.6291 \\
        LLM-as-a-Judge Std    & 0.0182 & 0.0098 \\
        \bottomrule
    \end{tabular}
    \caption{Accuracy scores (Mean and Std) across paraphrased prompts for older LLMs.}
    \label{tab:older-llms}
\end{table}

From Table~\ref{tab:older-llms}, we find that using LLM-as-a-Judge significantly reduces variance across paraphrased prompts. These additional results indicate that the reduction of prompt-induced variability is not due to recent models ``fixing'' the issue, but rather that such variability has consistently been an artifact of evaluation methods rather than a fundamental inconsistency in model behavior.

\end{document}